\documentclass[conference]{IEEEtran}

\usepackage{blindtext, graphicx}
\usepackage[backend=bibtex,style=ieee]{biblatex}
\usepackage[T1]{fontenc}

\ifCLASSOPTIONcompsoc
  \usepackage[caption=false,font=normalsize,labelfont=sf,textfont=sf]{subfig}
\else
  \usepackage[caption=false,font=footnotesize]{subfig}
\fi

\usepackage[cmintegrals]{newtxmath}

\usepackage{caption}
\usepackage{float}
\usepackage{amsmath}
\usepackage{varwidth}
\usepackage{float}

\usepackage[ruled,vlined,linesnumbered]{algorithm2e}
\usepackage{amsmath}
\usepackage[noend]{algpseudocode}
\IEEEoverridecommandlockouts

\makeatletter
\def\BState{\State\hskip-\ALG@thistlm}
\makeatother

\usepackage{tikz}   
\usepackage{siunitx} 
\usepackage{pgfplots}
    \usetikzlibrary{
        pgfplots.colorbrewer,
    }
    \pgfplotsset{
        cycle list/.define={my marks}{
            every mark/.append 
            style={solid,fill=\pgfkeysvalueof{/pgfplots/mark list fill}}\\
            every mark/.append style={solid,fill=\pgfkeysvalueof{/pgfplots/mark list fill}},mark=square*\\
            every mark/.append style={solid,fill=\pgfkeysvalueof{/pgfplots/mark list fill}},mark=triangle*\\
            every mark/.append style={solid,fill=\pgfkeysvalueof{/pgfplots/mark list fill}},mark=diamond*\\
            every mark/.append
            style={solid,fill=\pgfkeysvalueof{/pgfplots/mark list fill}},mark=*\\
        },
    }

\begin{document}

\title{Approximating the Void: Learning Stochastic Channel Models from Observation with Variational Generative Adversarial Networks}

\author{\IEEEauthorblockN{Timothy J. O'Shea}
\IEEEauthorblockA{DeepSig Inc \& Virginia Tech\\
Arlington, VA\\
toshea@deepsig.io}
\and
\IEEEauthorblockN{Tamoghna Roy}
\IEEEauthorblockA{DeepSig Inc.,\\
Arlington, VA\\
troy@deepsig.io}
\and
\IEEEauthorblockN{Nathan West}
\IEEEauthorblockA{DeepSig Inc.,\\
Arlington, VA\\
nwest@deepsig.io}
}

\maketitle
\begin{abstract}
Channel modeling is a critical topic when considering designing, learning, or evaluating the performance of any communications system.  Most prior work in designing or learning new modulation schemes has focused on using highly simplified analytic channel models such as additive white Gaussian noise (AWGN), Rayleigh fading channels or similar.  Recently, we proposed the usage of a generative adversarial networks (GANs) to jointly approximate a wireless channel response model (e.g. from real black box measurements) and optimize for an efficient modulation scheme over it using machine learning.  This approach worked to some degree, but was unable to produce accurate probability distribution functions (PDFs) representing the stochastic channel response.  In this paper, we focus specifically on the problem of accurately learning a channel PDF using a variational GAN, introducing an architecture and loss function which can accurately capture stochastic behavior.  We illustrate where our prior method failed and share results capturing the performance of such as system over a range of realistic channel distributions.
\end{abstract}


\IEEEpeerreviewmaketitle

\section{Introduction}
\label{sec:into}

\begin{figure}[!ht]
    \centering
    \includegraphics[width=0.5\textwidth]{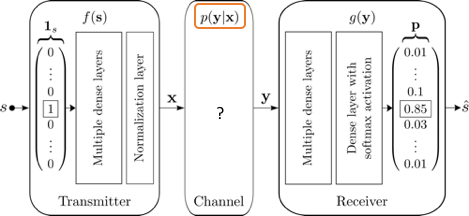}
    \caption{A channel autoencoder system for learning physical layer encoding schemes optimized for a stochastic channel model expression}
    \label{fig:cae}
\end{figure}

Recent work in machine learning based communications systems has shown that the autoencoder can be used very effectively to jointly design modulation schemes through parametric encoding and decoding networks under complex impairments while obtaining excellent performance \cite{intromlcomsys} (basic architecture shown in Figure \ref{fig:cae}).  

This approach generally employs a parametric encoder network $f(s,\theta_f)$ which encodes symbols $s$ into a transmitted symbol $x$, a stochastic channel model $y = h(x)$, and a decoder $g(y,\theta_g)$ which recovers estimates for the transmitted symbols $\hat{s}$ from the received samples $y$.  Both networks employ dense neural network style functions with weight and bias parameters $\theta_{f}$ and $\theta_{g}$ whose weight vectors may be optimized to learn many different non-linear mappings and function approximations.

The major drawback of this work however, is that it relies on having a differentiable channel model function ($\hat{y} = h(x)$), so that the gradients can be computed during back-propagation while training the network to minimize the reconstruction error rate (e.g. by directly computing the channel gradient $\frac{\partial h(x)}{\partial x}$ for use in the chain rule when computing the gradient of the loss with respect to the encoder network weights).  Without a differentiable channel model, the best we can do is optimize only-the decoder portion of the network, given ground truth label information, as discussed in \cite{dorner2017deep}.  The simplest forms of such an analytic channel model include the additive white Gaussian noise (AWGN) channel, but can also include a wide range of effects such as device non-linearities, propagation effects, fading, interference, or other distortions.   Figure \ref{fig:cae} contains a high level illustration of this training architecture.   Since the channel model, $h(x)$ is a stochastic function, it may also be represented as a conditional probability distribution $p(y|x)$, which more realistically approximates the random behavior of many channel phenomena.

In many applications it may be desirable to optimize performance for specific over the air channel effects or scenarios which exhibit several combined channel effects (e.g. devices response, interference effects, distortion effects, noise effects).  While simplified analytic models can be used in some cases to capture some of these effects, often real world non-linear effects and especially combinations of such effects are not well captured by these models simply due to their complexity and degrees of freedom which can be hard to capture in compact expressions.   In this work we focus on a more model-free approach for approximating stochastic channel responses, allowing for high degrees of freedom, by using variational neural networks \cite{kingma2013auto} in order to approximate the end-to-end responses based on real world measurement data.
This approach is appealing in its degree of flexibility, accuracy, and the degree with which such a channel approximation network can be used to directly optimize the modulation and encoding methods for a corresponding communications system to attain excellent real world performance under a wide range of conditions, effects and constraints.

To accomplish this, we consider the task of jointly approximating the wireless channel response and a good encoding therefore using a generative adversarial network (GAN) \cite{goodfellow2014generative} in our prior work \cite{o2018physical}.  Here, we leverage three neural networks, an encoder $f(s,\theta_f)$, a channel approximation $h(x,\theta_h)$, and a decoder $g(y,\theta_g)$.  Each of these is comprised principally of fully connected (FC) rectified linear unit (ReLU) layers \cite{nair2010rectified}, where  the transfer function for a single FC-RELU layer is shown below in equation \ref{eq:fcrelu}.  Here, the layer output values $\vec{\ell+1}$, are computed from the input values $\vec{\ell}$, weight vector $\vec{w}$, bias vector $\vec{b}$, while using a rectifier as a non-linearity.  Where $i$ denotes the layer index, $k$ denotes the input index, and $j$ denotes the output index.  An equivalent FC layer without the RELU activation is used at the output of the encoder, and channel approximation layers as shown in equations \ref{eq:fc}.  Each network's parameters $\theta$ then constitute the weight $\vec{w}$ and bias $\vec{b}$ values for each of the network's layers.

\begin{equation}
    \label{eq:fcrelu}
    \vec{\ell}_{i+1,j} = max(0, \sum_k \vec{\ell}_{i,k} \vec{w}_{i,k} + \vec{b}_{i,j})
\end{equation}

\begin{equation}
    \label{eq:fc}
    \vec{\ell}_{i+1,j} = \sum_k \vec{\ell}_{i,k} \vec{w}_{i,k} + \vec{b}_{i,j}
\end{equation}

A variety of loss functions can be used to optimize these networks, in the prior work, mean squared error (MSE) was used to to perform regression of transmit and received sample values ($x$ and $y$) to approximate the channel response, wherein network parameters are updated using stochastic gradient of the MSE loss function as shown in equation \ref{eq:loss}.

\begin{equation}
    \label{eq:loss}
    \nabla_{\theta_h} \left(y - h(x,\theta_h)\right)^2
\end{equation}

In this work, we investigate instead treating the channel network $y = h(x)$ as a stochastic function approximation and replicate the resulting conditional probability distribution $p(y|x)$ instead, which yields a significantly more appropriate tool for optimization, test, and measurement of communications systems.   

\section{Technical Approach}

Generative adversarial networks (GANs), introduced in \cite{goodfellow2014generative} are a powerful class of generative models that consist of two or more competing objective functions which help reinforce each other.  Typically this consists of a generator, which translates a latent space into a high dimensional sample which mimics some real data distributions, and a discriminator which attempts to classify samples as real or fake (i.e. produced by the generator network).  By jointly/iteratively training these two networks, and by leveraging the backwards pass gradient of the discriminator to train the generator towards the "true" label, this approach has proven to be extremely effective in producing synthetic generators which produce data samples indistinguishable from real dataset samples for a range of common visual tasks.  Numerous improvements have been made to the approach since its introduction, including variational network features, training stability improvements, architecture enhancements and others further explored in \cite{srivastava2017veegan,CGAN,DCGAN,WGAN,InfoGAN}.  

Most of these works focus on computer vision applications of GANs, for instance for image generation.  We focus in contrast on the ability of variational-GANs to approximate robust probability distributions for the task of approximating accurate conditional channel distribution $p(y|x)$ primarily for the purpose of training autoencoder based communications systems as described in section \ref{sec:into}.



We consider the channel approximation network $\hat{y} = h(x,\theta_h)$ to be a conditional probability distribution, $p(\hat{y}|x)$ and instead of minimizing $\mathcal{L}_{MSE}(y,\hat{y})$ as before, we seek to minimize the distance between the conditional probability distributions $p(y|x)$ and $p(\hat{y}|x)$ resulting from measurement and from the variational channel approximation network respectively.  This can be accomplished as in \cite{goodfellow2014generative} by minimizing parameters of each network using the two stochastic gradients given below in equations \ref{eq:bceloss1} and \ref{eq:bceloss2} where we introduce a new discriminative network $D(x_i,y,\theta_D)$ to classify between real ($y$) and synthetic samples ($\hat{y}$) from the channel given its input ($x$).  In this case, $h(x,\theta_h)$ takes the place of the generative network (often written as $G(z)$), where $x$ reflects conditional transmitted symbols/samples, and additional stochasticity in the function is introduced by variational layers.

\begin{equation}
    \label{eq:bceloss1}
    \nabla_{\theta_D} \frac{1}{N} \sum_{i=0}^{N} \left[ \text{log} \left( D(x_i,y_i,\theta_D) \right) + \text{log}\left( 1 - D(x_i,h(x_i,\theta_h),\theta_D)  \right) \right]
\end{equation}
\begin{equation}
    \label{eq:bceloss2}
    \nabla_{\theta_h} \frac{1}{N} \sum_{i=0}^{N} \text{log} \left( 1 - D(x_i,h(x_i,\theta_h), \theta_D) \right)
\end{equation}

This optimization can also be performed using a Wasserstein GAN \cite{WGAN} to improve training stability wherein updates can be made according to the stochastic gradient given in equations \ref{eq:wgan1} and \ref{eq:wgan2}.

\begin{equation}
    \label{eq:wgan1}
    \nabla_{\theta_D} \frac{1}{N} \sum_{i=0}^{N} \left[ \left( D(x_i,y_i,\theta_D) \right) - D(x_i,h(x_i,\theta_h),\theta_D)  \right]
\end{equation}
\begin{equation}
    \label{eq:wgan2}
    \nabla_{\theta_h} \frac{1}{N} \sum_{i=0}^{N} D(x_i,h(x_i,\theta_h),\theta_D) 
\end{equation}

Within the channel approximation network, the variational sampler layer samples from a random distribution (the Gaussian distribution in this case) parameterized by the outputs of the previous layer.  The number of FC ReLU and Linear(LIN) layers is not-tuned in this case, but should be wide and deep enough to express the complexity of the topology required for the mapping.  This may vary for different applications and should be tuned as with any architecture or set of hyper-parameters for your application.  Linear layers are used for regression of all real values for sampler parameters and for network output, the full architecture is shown in table \ref{table:hnet} for the channel approximation network and in table \ref{table:Dnet} for the discriminative network.

\begin{equation}
    \label{eq:var}
    \vec{\ell}_{i+1,j} = \vec{N}(\mu=\vec{\ell}_{i,2j},\sigma=\vec{\ell}_{i,2j+1}) 
\end{equation}

\begin{table}
\begin{center}
\begin{tabular}{ c c c }
 Layer & Outputs & Params in $\theta_h$ \\ 
 \hline
 FC-RELU & 20 & $w_0,b_0$ \\
 FC-RELU & 20 & $w_1,b_1$\\
 FC-RELU & 20 & $w_2,b_2$\\
 FC-LIN & 32  & $w_3,b_3$\\
 Sampler & 16 & None \\
 FC-RELU & 80 & $w_4,b_4$\\
 FC-LIN & 2   & $w_5,b_5$ \\
\end{tabular}
\end{center}
\caption{Channel Approximation Network, $h(x,\theta_h)$}  
\label{table:hnet}
\end{table}

\begin{table}
\begin{center}
\begin{tabular}{ c c c }
 Layer & Outputs & Params in $\theta_D$ \\ 
 \hline
 FC-RELU & 80 & $w_0,b_0$ \\
 FC-RELU & 80 & $w_1,b_1$\\
 FC-RELU & 80 & $w_2,b_2$\\
 FC-Sigmoid & 1  & $w_3,b_3$\\
\end{tabular}
\end{center}
\caption{Channel Discriminative Network, $D(x, y,\theta_D)$}  
\label{table:Dnet}
\end{table}

Optimization of these networks is performed iteratively (e.g. one mini-batch of each, alternating between objective functions and update parameter sets) and using the Adam \cite{kingma2014adam} optimizer with a learning rate between 1e-4 and 5e-4.

\begin{figure}[!ht]
    \centering
    \includegraphics[width=0.5\textwidth]{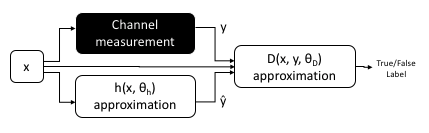}
    \caption{High level training architecture for conditional-variational-GAN based learning of stochastic channel approximation function}
    \label{fig:gan}
\end{figure}

\section{Measurements and Results}

Considering one of the simplest canonical communications system formulations, we focus first on the case of a binary phase-shift keying (BPSK)-AWGN system where, $p(x)$ is a discrete uniform (IID) random variable over all (2) possible symbol values (e.g. whitened information bits), in this case $x \in \{ -1,+1 \}$ and $p(x) = [0.5,0.5]$.  This encoding scheme, while fixed in this work, can easily be updated as an additional optimization network process using a channel autoencoder as described in \cite{intromlcomsys,o2018physical} over the learned stochastic channel approximation function.  We first consider the AWGN channel $h(x)$ given by $h(x) = x + \text{N}(0,1.0)$.  

\begin{figure}[!ht]
    \centering
    \includegraphics[width=0.5\textwidth]{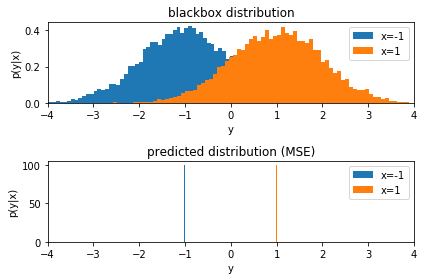}
    \caption{Learned distribution using direct MSE minimization}
    \label{fig:dist1}
\end{figure}

If we consider the direct approach to training a [non-stochastic] channel function approximation (e.g. no variational layer), using MSE less directly as expressed in equation \ref{eq:loss}, we obtain the two resulting distributions shown in figure \ref{fig:dist1}.  Here, the blackbox distribution results from measurement of the channel based on the ground truth stochastic process, while the predicted distribution reflects the expected behavior of the channel approximation network.



\begin{figure}[!ht]
    \centering
    \includegraphics[width=0.4\textwidth]{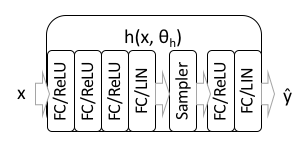}
    \caption{Variational architecture for the stochastic channel approximation network (conditional generator)}
    \label{fig:varchan}
\end{figure}

We can see here that the channel network approximation when trained with MSE loss rapidly converges to the same mean values for each conditional input value.   Unfortunately, we do not accurately learn an appropriate variance or full distribution to reflect the channel effects at all using this approach.  Our conditional generator $h(x,\theta_h)$ in this case is a deterministic function (not including the variational layer, but also when training the variational layer using only MSE loss).  It can not accurately reflect this mapping from the discrete valued $x$ distribution to real continuous distributions over $y$ without the variational layer.  To this end, we instead consider the channel approximation function $h(x,\theta_h)$ \textbf{with} the variational sampling layer, with the architecture shown in Figure \ref{fig:varchan}, where a latent space $z$ is sampled from latent distribution parameters $\theta_z$ produced by the first FC/LIN layer within the hidden layers of the network.  

\begin{figure}[!ht]
    \centering
    \includegraphics[width=0.5\textwidth]{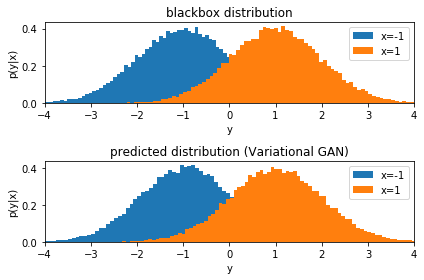}
    \caption{Learned distribution using variational GAN training}
    \label{fig:dist2}
\end{figure}

In contrast, if we \textbf{do} include the variational layer, and consider the GAN based training approach (using cross-entropy loss and a discriminator network), we can obtain significantly better performance.  Figure \ref{fig:dist2} illustrates the measured and approximated conditional distributions for the same channel when using this approach.  We see that the distributions match well in both mean and variance this time, closely resembling the appropriate BPSK-AWGN channel response which has been measured.  In this way, we've learned a model which can accurately reflect the stochastic channel behavior, and produce be used for training or tuning a communications system which closely matches the performance of the real world as far as the distribution $p(y|x)$ is concerned, for instance reflecting a nearly identical signal-to-noise ratio in this case. 

\begin{figure}[!ht]
    \centering
    \includegraphics[width=0.5\textwidth]{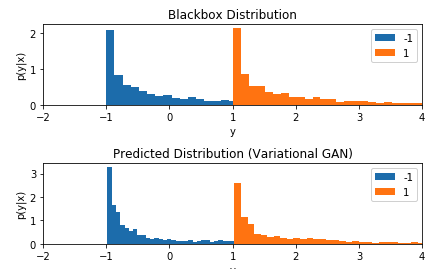}
    \caption{Learned 1D distributions of conditional density on non-Gaussian (Chi-Squared) channel effects using variational GAN training}
    \label{fig:dist_chi}
\end{figure}

The appeal of this approach however, is its applicability beyond a simple AWGN channel to much more complex channel effects which are unconventional, non-linear and hard-to-express in compact parametric analytic forms.  We can consider the case for instance of a similar BPSK style communications system, with an additive Chi-Squared channel distribution.   Using the same variational GAN based approach for conditional PDF channel approximation, we can still rapidly converge on a representative non-Gaussian distribution formed from the same channel approximation network.  In this case, the 16 latent variables sampled from the learned latent parameter space are all Gaussian, but they combine to form an additive approximation for the non-Gaussian distribution.  Measured and approximated conditional distributions from the black box channel model are shown in Figure \ref{fig:dist_chi}.  There is definitely some error present in the resulting distribution from this approximation, resulting in part from its representation as a mixture of Gaussian latent variables, but this can be alleviated by choosing different sampling distributions and by increasing the dimension size of the latent space (at the cost of increased model complexity).

\subsection{Scaling Dimensionality}

\begin{figure}[!ht]
    \centering
    \includegraphics[width=0.5\textwidth]{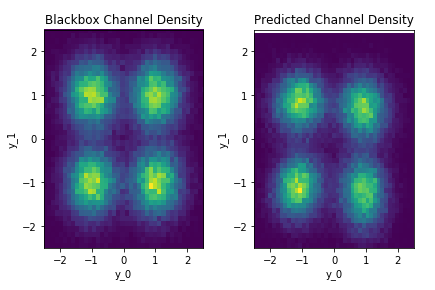}
    \caption{Learned 2D distribution for 4-QAM using variational GAN training on AWGN data}
    \label{fig:dist2d1}
\end{figure}

\begin{figure}[!ht]
    \centering
    \includegraphics[width=0.5\textwidth]{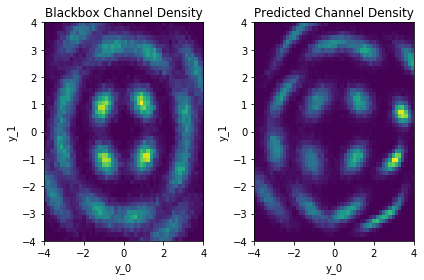}
    \caption{Learned 2D distributions of received 16-QAM constellation non-linear channel effects using variational GAN training}
    \label{fig:dist2d2}
\end{figure}

This approach can be readily scaled to higher dimensional channel responses, as well as complex cascades of stochastic effects, jointly approximating the aggregate distribution with the network.  We expand the real-valued BPSK case to a canonical complex quadrature representation of symbols using in-phase and quadrature basis functions.  Figure \ref{fig:dist2d2} illustrates a heat-map for the measured and approximated probability density functions ($p(y|x)$) for a QPSK system with a simple AWGN channel.   In this plot, we marginalize over the conditional $x$ and show simply $ p(y) = \sum_{i=0}^N \frac{1}{N} p(y|x_i)$.  

While the channel in this case is still quite simple, we can see  that the variational GAN has readily learned appropriate statistics for the distribution, which matches the measured distribution accurately.

While there were simple examples, the real appeal of this approach is its ability to scale to new modulation types, new channel effects, accurate non-linear models, and new complex combinations of channel impairments which may be present in any given black box system or measurement campaign.  To illustrate this we include a final experiment with a higher degree of realism, modeling a 16-QAM system which includes AWGN effects as well as phase noise, phase offset, and non-linear AM/AM and AM/PM effects introduced by a hardware amplifier model.  There are all impairments which are typically encountered in a wide range of devices and communications systems.  Figure \ref{fig:dist2d2} illustrates as with the previous examples, the marginalized $p(x)$ distribution for both the measured version of the received signal, and the approximated version of the distribution when learning a stochastic channel approximation model with our approach.  In this case, we can see a number of interesting effects are learned, including each constellation point's distribution, circumferential elongation of these distributions due to phase noise at higher amplitudes, and generally a decent looking first order approximation of the distribution.  

The distribution is notably not perfect in this example, we can see some differences and errors in the estimation of the radial variance of these distributions, especially at higher amplitudes.  We believe this can be addressed through improved model training, for instance using the WGAN approach to improve stability, along with the use of larger models with more degrees of freedom to facilitate accurate representation.  Numerous additional architecture enhancements are possible, for instance ones that could take advantage of the polar nature of the representation, which may simplify the representation of some of the effects what are often modeled in polar terms more simply.  Time varying behaviors such as SNR variation, fading, etc can be easily extended from this model by adding temporal dependencies into the network for instance with series of channel samples, and RNN-style sequence modeling of density approximations in the model. 

\section{Discussion} 

Channel modeling has always been a difficult but critical task within wireless communications systems.  With accurate stochastic models of a wireless channel, we can design and optimize communications systems for them using both closed form analytic modeling approaches and higher more scalable machine learning based approaches such as the channel autoencoder.  By adopting a mostly model-free learning approach to channel modeling using deep learning and variational-GANs, we illustrate in this work that a range of different types of stochastic channel models can be learned accurately from measurement, without the introduction of many assumptions about the effects occurring, or the simplification to a parametric model.  

This approach holds to potential then to accurately reflect a wide range of stochastic channel behaviors, provides a convenient, compact, and uniform way to represent them, allows for high rate sampling and simulation from these models, and lends itself to scaling such models to very high degrees of freedom.  In doing so, we hope that this approach will scale well to systems with numerous hardware effects, hardware impairments, multiple-antennas, and any other sort of stochastic impairment which may be measured within a communications system.  Such a model can then be readily paired with an autoencoder based approach and used either pre-trained, or during joint training to optimize communications systems and new modulation types directly for many real world deployment scenarios in a highly generalizable way, with little specific manual optimization or specification needed.

While analytic closed form parametric channel models and understanding of stochastic wireless impairments will always be important in modeling and thinking about wireless systems, this basic approach to channel approximation offers an important tool in dealing with complexity and degrees of freedom.  By making less assumptions and by more accurately modeling end-to-end system behavior in a comprehensive way, we believe this will lead to better models and better performance for an entire class of future wireless systems.


Further more, by providing accurate stochastic differentiable approximations of these complex aggregation of propagation effects, we can readily optimize encoding and decoding schemes on both ends of a link (using backpropagation) to achieve near optimal performance metrics.  This is one the collection of methods DeepSig is leveraging in order to train, validate, and adapt their prototype next generation learned physical layer communications systems for specific channels in over the air and unique deployment configurations. 

Such over the air, data-centric learning methods in communications systems stand to become increasingly important in the future as as multi-antenna and multi-user and many device systems continue to increase in complexity and plurality of possible configurations and deployment scenarios, each benefiting from a slightly different tailored set of physical layer adaptations.  While many enhancements may exist for this approach in terms of improving GAN stability and performance, this work illustrates that such an approach can, even with relatively simple variational GANs, obtain reasonably accurate channel approximation performance for common wireless channel models and effects.   Significantly more work remains to be done in this field



\printbibliography
\end{document}